\documentclass[sigconf]{acmart}

\usepackage{multirow}
\usepackage{overpic}
\usepackage{float}
\usepackage{graphicx}
\usepackage{subfig}

\newcommand{\etal}{\textit{et al}.}
\newcommand{\ie}{\textit{i}.\textit{e}.}
\newcommand{\eg}{\textit{e}.\textit{g}.}
\newcommand{\etc}{\textit{etc}.}

\AtBeginDocument{%
  }

\setcopyright{acmlicensed}
\copyrightyear{2024}
\acmYear{2024}
\acmDOI{10.1145/3664647.3680868}

\acmConference[MM '24] {Proceedings of the 32nd ACM International Conference on Multimedia}{October 28--November 1, 2024}{Melbourne, VIC, Australia.}
\acmBooktitle{Proceedings of the 32nd ACM International Conference on Multimedia (MM '24), October 28--November 1, 2024, Melbourne, VIC, Australia}
\acmISBN{979-8-4007-0686-8/24/10}

\begin{document}

\title{Subjective-Aligned Dataset and Metric for Text-to-Video Quality Assessment}

\author{Tengchuan Kou}
\affiliation{%
 \institution{Shanghai Jiao Tong University}
 \city{Shanghai}
 \country{China}
 \orcid{0009-0001-1510-027X}
}
\email{2213889087@sjtu.edu.cn}

\author{Xiaohong Liu}
\authornotemark[1]
\affiliation{%
 \institution{Shanghai Jiao Tong University}
 \city{Shanghai}
 \country{China}}
\email{xiaohongliu@sjtu.edu.cn}

\author{Zicheng Zhang}
\affiliation{%
 \institution{Shanghai Jiao Tong University}
 \city{Shanghai}
 \country{China}}
\email{zzc1998@sjtu.edu.cn}

\author{Chunyi Li}
\affiliation{%
  \institution{Shanghai Jiao Tong University}
  \city{Shanghai}
  \country{China}
}
\email{lcysyzxdxc@sjtu.edu.cn}

\author{Haoning Wu}
\affiliation{%
 \institution{Nanyang Technological University}
 \city{Singapore}
 \country{Singapore}}
\email{haoning001@e.ntu.edu.sg}

\author{Xiongkuo Min}
\affiliation{%
 \institution{Shanghai Jiao Tong University}
 \city{Shanghai}
 \country{China}}
\email{minxiongkuo@sjtu.edu.cn}

\author{Guangtao Zhai}
\affiliation{%
 \institution{Shanghai Jiao Tong University}
 \city{Shanghai}
 \country{China}}
\email{zhaiguangtao@sjtu.edu.cn}

\author{Ning Liu}
\authornote{Corresponding authors.}
\affiliation{%
 \institution{Shanghai Jiao Tong University}
 \city{Shanghai}
 \country{China}}
\email{ningliu@sjtu.edu.cn}
\renewcommand{\shortauthors}{Tengchuan Kou et al.}

\begin{abstract}
  With the rapid development of generative models, AI-Generated Content (AIGC) has exponentially increased in daily lives. Among them, Text-to-Video (T2V) generation has received widespread attention. Though many T2V models have been released for generating high perceptual quality videos, there is still lack of a method to evaluate the quality of these videos quantitatively. To solve this issue, we establish the largest-scale Text-to-Video Quality Assessment DataBase (\textbf{T2VQA-DB}) to date. The dataset is composed of 10,000 videos generated by 9 different T2V models, along with each video's corresponding mean opinion score. 
  Based on T2VQA-DB, we propose a novel transformer-based model for subjective-aligned Text-to-Video Quality Assessment (\textbf{T2VQA}). 
  The model extracts features from text-video alignment and video fidelity perspectives, then it leverages the ability of a large language model to give the prediction score. Experimental results show that T2VQA outperforms existing T2V metrics and SOTA video quality assessment models. Quantitative analysis indicates that T2VQA is capable of giving subjective-align predictions, validating its effectiveness. The dataset and code are available at \href{https://github.com/QMME/T2VQA}{https://github.com/QMME/T2VQA}. 
\end{abstract}


\begin{teaserfigure}
  \centering
  \includegraphics[width=0.96\textwidth]{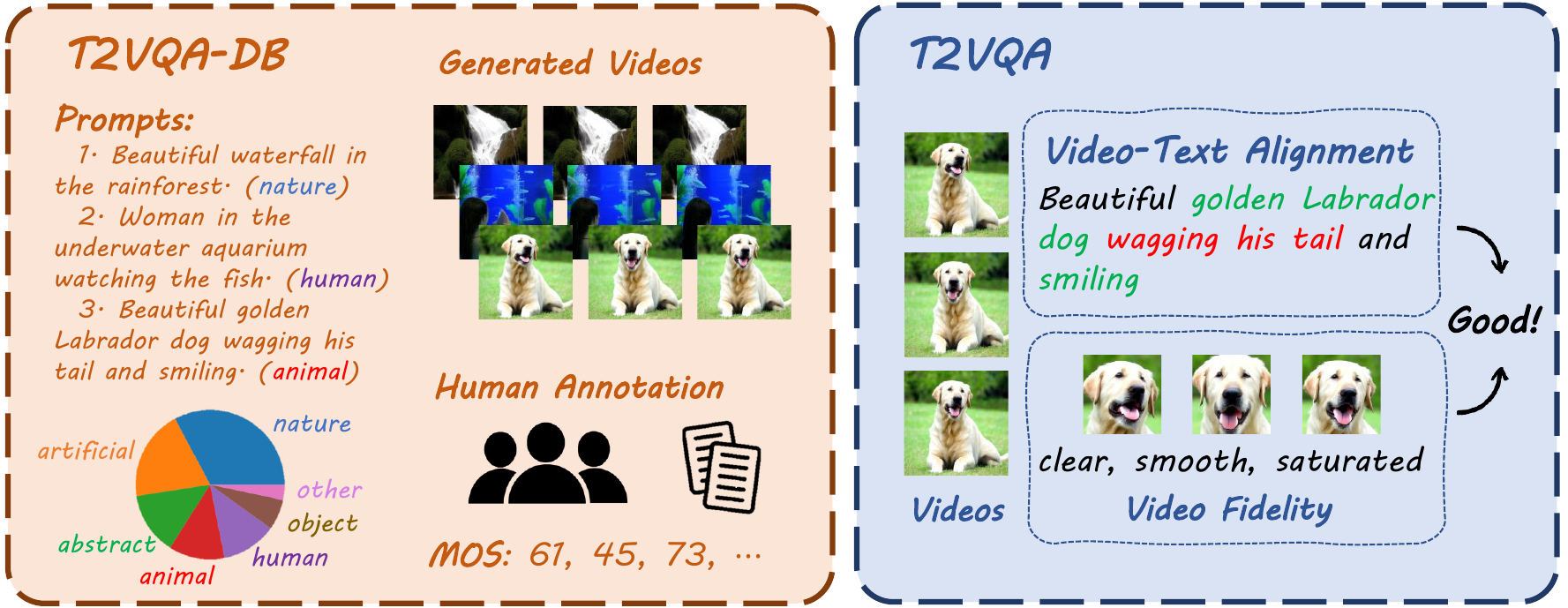}
  \caption{Overview of the proposed \textit{T2VQA-DB} and \textit{T2VQA}. T2VQA-DB has the largest scale among existing T2V datasets. T2VQA achieves the SOTA performance in evaluating the quality of text-generated videos.}
  \label{fig:overview}
\end{teaserfigure}


\maketitle

\section{Introduction}

\begin{figure*}[t]
    \captionsetup[subfloat]{labelformat=empty}       
		\centering

            \subfloat[t][Text2Video-Zero~\cite{khachatryan2023text2video}]{\includegraphics[height=0.133\linewidth]{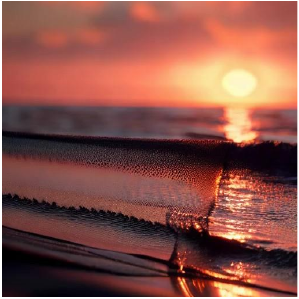}}
            \hspace{1.5mm}
            \subfloat[t][AnimateDiff~\cite{guo2023animatediff}]{\includegraphics[height=0.133\linewidth]{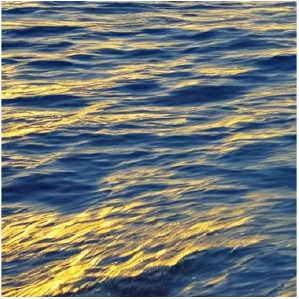}}
            \hspace{1.5mm}
            \subfloat[t][VidRD~\cite{gu2023reuse}]{\includegraphics[height=0.133\linewidth]{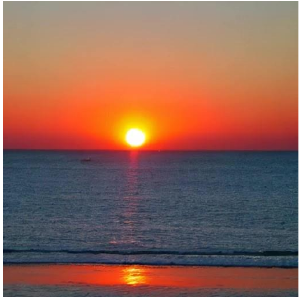}} 
            \hfill
            \subfloat[t][AnimateDiff~\cite{guo2023animatediff}. MOS: 62.62.]{\includegraphics[width=0.555\linewidth]{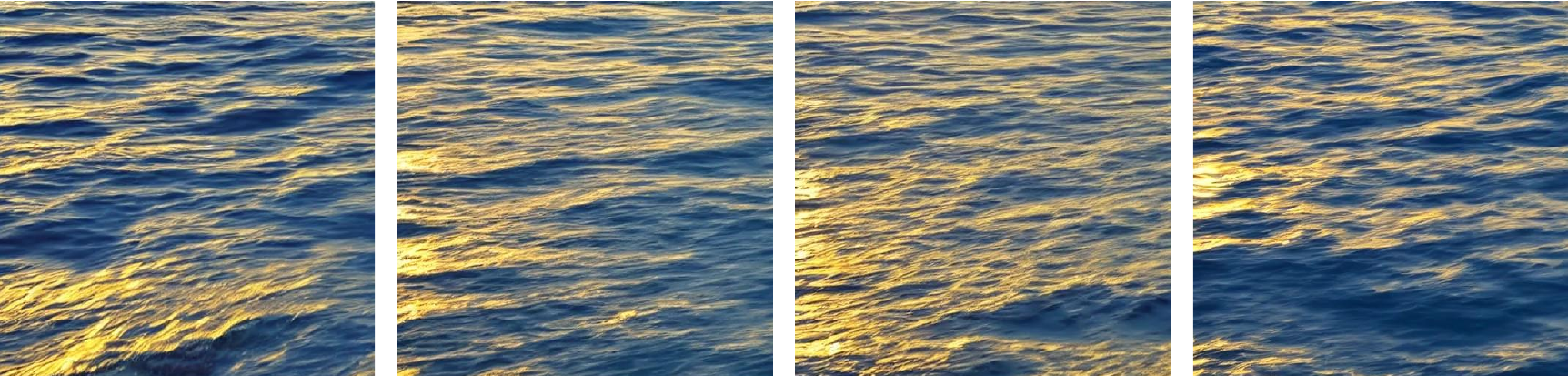}}
        \vspace{-2mm}
        \begin{minipage}[t]{\linewidth}
            \subfloat[][Tune-a-video~\cite{wu2023tune}]{\includegraphics[height=0.133\linewidth]{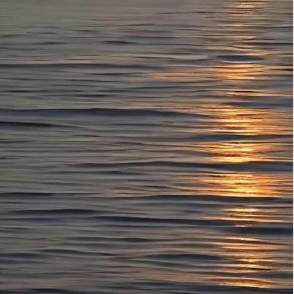}}
            \hspace{1.5mm}
            \subfloat[][VideoFusion~\cite{luo2023videofusion}]{\includegraphics[height=0.133\linewidth]{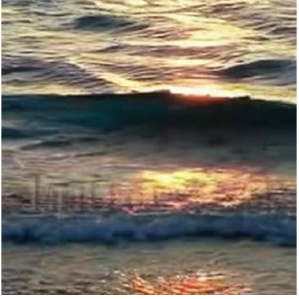}}
            \hspace{1.5mm}
            \subfloat[][LVDM~\cite{he2022latent}]{\includegraphics[height=0.133\linewidth]{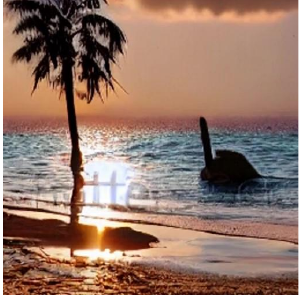}} 
            \hfill
            \subfloat[][LVDM~\cite{he2022latent}. MOS: 53.78]{\includegraphics[width=0.555\linewidth]{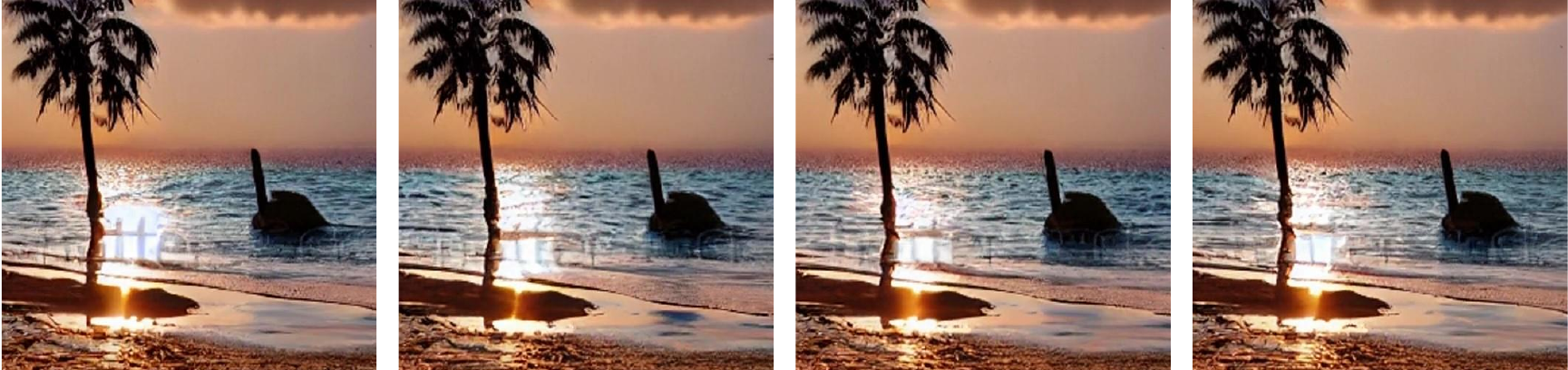}}
        \vspace{-2mm}    
        \end{minipage}
        
        \begin{minipage}[t]{\linewidth}
            \subfloat[][ModelScope~\cite{wang2023modelscope}]{\includegraphics[width=0.133\linewidth]{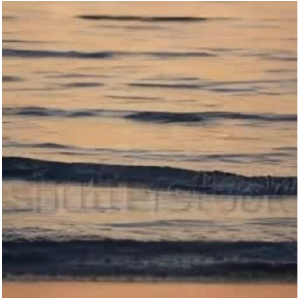}}
            \hspace{1.5mm}
            \subfloat[][Show-1~\cite{zhang2023show}]{\includegraphics[width=0.133\linewidth]{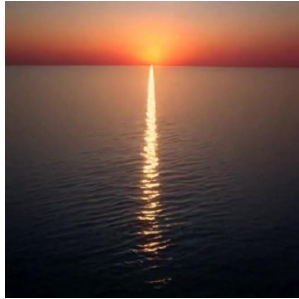}}
            \hspace{1.5mm}
            \subfloat[][LaVie~\cite{wang2023lavie}]{\includegraphics[width=0.133\linewidth]{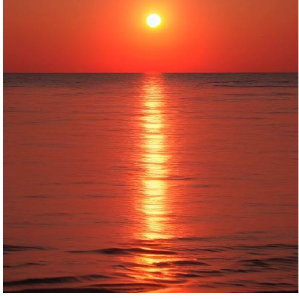}} 
            \hfill
            \subfloat[][LaVie~\cite{wang2023lavie}. MOS: 77.57]{\includegraphics[width=0.555\linewidth]{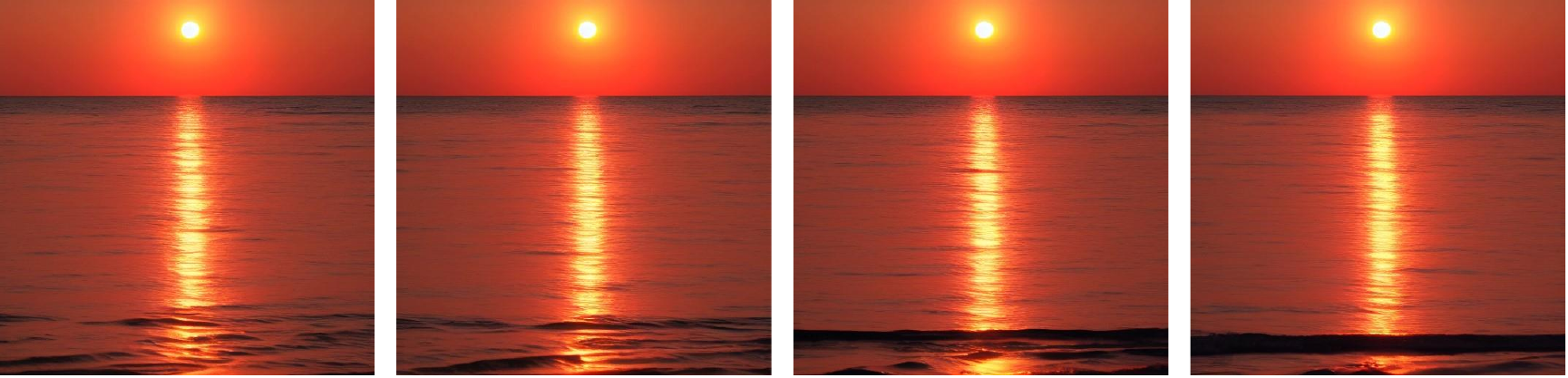}}
            \put(-400, -22){\small \textbf{(a)}}
            \put(-144, -22){\small \textbf{(b)}}
        \end{minipage}

    \vspace{-2mm}
    \caption{Video examples generated by prompt: \textit{Sunset on the sea}. (a) Overview of video frames generated by 9 models. (b) Videos generated by AnimateDiff~\cite{guo2023animatediff}, LVDM~\cite{he2022latent}, LaVie~\cite{wang2023lavie}, and their MOSs. }
    \label{fig:videos}
\end{figure*}

Video generation, or video synthesis, has been fully developed in the past few years. Text-to-Video (T2V) generation is one of the most studied fields, where a user provides a text description as the guidance for video generation. With the thriving of diffusion-based models, high-fidelity videos can be generated. However, the quality of text-generated videos is diverse which affects the experience quality of subjects. Therefore, a subjective-aligned quality assessment method for them is needed. Unfortunately, existing Video Quality Assessment (VQA) models are unable to accomplish the task well. On the one hand, distortions brought by T2V generation models, such as jitter effect, irrational objects, are different from distortions in natural videos. On the other hand, traditional VQA models do not take text-video alignment into consideration, which is a significant evaluation perspective for text-generated videos.

Besides, the most used metrics for T2V generation, such as IS~\cite{salimans2016improved}, FVD~\cite{unterthiner2018towards}, and CLIPSim~\cite{wu2021godiva}, fail to reflect real user preferences. IS uses the Inception Network~\cite{szegedy2016rethinking} to generate a distribution that reflects image/video quality and diversity. It has been criticized for its inability to evaluate image/video quality precisely.
FVD compares the I3D feature~\cite{carreira2017quo} distributions of the generated and natural video pair. 
The drawback of FVD is that obtaining the reference natural video is usually impractical. CLIPSim takes advantage of CLIP~\cite{radford2021learning} to calculate the similarity between the original text and the generated video content. However, it only considers text-video alignment from the image level, excluding the temporal information and perceptual video quality.

To facilitate the development of a more comprehensive and accurate metric, we establish the largest-scale subjective T2V dataset to date, named Text-to-Video Quality Assessment DataBase (\textbf{\textit{T2VQA-DB}}). The dataset contains 10,000 videos generated by 9 representative T2V models using 1,000 text prompts. We also collect each video's Mean Opinion Score (MOS) by conducting a subjective experiment, where 27 subjects score the overall quality of the generated videos. 
Fig.~\ref{fig:videos} shows video frames generated by the 9 T2V models from the prompt ``Sunset on the sea''. And we give more detailed overviews of three examples. We anticipate that the T2VQA-DB will benefit the development of subsequent models.

Based on T2VQA-DB, we propose a novel model equipped with multi-modality foundation models for better Text-to-Video Quality Assessment (\textbf{\textit{T2VQA}}). 
The model utilizes BLIP~\cite{li2022blip} and Video Swin Transformer (Swin-T)~\cite{liu2022video} to extract features from text-video alignment and video fidelity perspectives respectively. The features are fused through a cross-attention module, then they are fed into a frozen Large Language Model (LLM) to regress the predicted score. We train and test T2VQA as well as other VQA models on T2VQA-DB. Experimental results show that T2VQA outperforms existing T2V generation metrics and state-of-the-art VQA models, validating its effectiveness in measuring the perceptual quality of text-generated videos. Fig.~\ref{fig:overview} shows the overview of the proposed T2VQA-DB and T2VQA.

We summarize our contributions as follows:

\begin{enumerate}
    \item We establish the T2V dataset with \textit{the largest scale to date}, named T2VQA-DB, which includes 10,000 text-generated video sequences and their corresponding MOSs gained from 27 subjects. 
    \item We propose a novel transformer-based model for better evaluating the quality of text-generated videos, named T2VQA. The model dissolves the problem from \textit{text-video alignment} and \textit{video fidelity} perspectives, and then it leverages the ability of an LLM to give a subjective-aligned prediction of the video quality. 
    \item The proposed T2VQA outperforms existing T2V generation metrics and SOTA VQA models on T2VQA-DB and other T2V datasets, indicating the effectiveness of T2VQA. Qualitative experiments show that T2VQA can benefit in measuring the performance of T2V generation algorithms, giving it practical application prospects.
\end{enumerate}

\begin{table}[]
  \caption{Summary of T2V datasets. [Bold: the best].}
  \label{tab:dataset}
  \centering
  \begin{tabular*}{\linewidth}{@{\extracolsep{\fill}}ccccc@{}}
    \toprule
    Name & Videos & Prompts & Models & Annotators \\
    \midrule
    Chivileva's~\cite{chivileva2023measuring} & 1,005 & 201 & 5 & 24\\
    ECTV~\cite{liu2023evalcrafter} & 5,600 & 700 & 8 & 7 \\
    VBench~\cite{huang2024vbench} & 6,984 & \textbf{1,746} & 4 & - \\
    FETV~\cite{liu2024fetv} & 2,476 & 619 & 4 & 3 \\
    TVGE~\cite{wu2024better} & 2,543 & 512 & 5 & 10\\
    \textbf{T2VQA-DB (ours)} & \textbf{10,000} & 1,000 & \textbf{9} & \textbf{27} \\
  \bottomrule
  \end{tabular*}
  \vspace{-2mm}
\end{table}

\section{Related Works}

\subsection{Text-to-video Dataset}

To the best of our knowledge, only a few T2V datasets have been proposed. They mainly have the following two issues: (1) Insufficient scale: Chivileva~\etal~\cite{chivileva2023measuring} propose a dataset with 1,005 videos generated by 5 T2V models following~\cite{li2023agiqa}. EvalCrafter~\cite{liu2023evalcrafter} builds ECTV using 700 prompts and 8 T2V models, resulting in 5,600 videos in total. Similarly, FETV~\cite{liu2024fetv} is composed of 2,476 videos generated by 619 prompts, and 4 T2V models. VBench~\cite{huang2024vbench} has a larger scale with in total of $\sim$1.7k prompts and 4 T2V models. TVGE~\cite{wu2024better} uses the prompts from ECTV and uses 5 T2V model for video generation. Such scales are not sufficient for the training of deep learning-based models, and cannot represent current T2V algorithms. (2) Limited human annotation: ITU-standard~\cite{bt2002methodology} requires at least 15 human annotators for subjective study. The dataset proposed by Chivileva~\etal is the only one that meets the standard. ECTV and FETV only have 3 users for annotation. VBench~\cite{huang2024vbench} does not specify the number of human annotators.


\subsection{Metrics for T2V Generation}

IS~\cite{salimans2016improved} and FVD~\cite{unterthiner2018towards} are the two most commonly used metrics for evaluating the quality of generated videos. IS uses the Inception feature to present both image/video quality. FVD measures the distance between the generated video and the natural video. However, both metrics are criticized for poor correlation with human visual perception. CLIPSim~\cite{wu2021godiva} measures the text-video alignment by using CLIP~\cite{radford2021learning}. After measuring the similarity between the text and each video frame, it averages them to get the final score. As a result, it only evaluates videos from the image level, losing the information from the temporal domain. As well as it doesn't consider the video quality. 

Though many works targeting the evaluation of text-generated images have been proposed~\cite{xu2024imagereward, kirstain2024pick, wu2023human, li2023agiqa, li2024q, zhang2023perceptual, li2024aigiqa, yang2024aigcoiqa2024}, there are only a few metrics tailored for text-generated video evaluation have been proposed. Among them, ViCLIP~\cite{wang2023internvid} is a CLIP-based metric for measuring text-video alignment. Chivileva~\etal~\cite{chivileva2023measuring} proposes an ensemble video quality metric that integrates text similarity and naturalness. EvalCrafter~\cite{liu2023evalcrafter} and VBench~\cite{huang2024vbench} build benchmarks to evaluate text-generated video from 18 and 16 objective metrics respectively. FETV~\cite{liu2024fetv} and T2V-Score~\cite{wu2024better} both propose separate metrics for text-video alignment and video quality, without an overall perceptual score for text-generated videos. There is still lack of a simple and effective metric to evaluate the quality of text-generated videos.

Besides the aforementioned metrics, VQA models can also be used for the evaluation of text-generated videos. BVQA~\cite{li2022blindly} transfers knowledge from Image Quality Assessment (IQA) databases and then trains on the target VQA database, \eg, KoNVid-1k\cite{hosu2017konstanz}. SimpleVQA~\cite{sun2022deep} extracts spatial and motion features to regress to the final score. FAST-VQA~\cite{wu2022fast} proposes ``fragments'' as a novel sampling strategy and the fragment attention network to accommodate fragments as inputs. ~\cite{gao2023vdpve, dong2023light, kou2023stablevqa, zhang2023advancing, zhang2024reduced, zhou2024light,zhou2024thqa} are works for the evaluation of enhanced videos and digital humans. DOVER~\cite{wu2023dover} proposes to view the quality assessment problem from technical and aesthetic perspectives, while BVQI~\cite{wu2023bvqi} and MaxVQA~\cite{wu2023explainable} integrate text prompts (\eg, good, bad) into VQA. With the development of Multi-modal Large Language Models (MLLMs), researchers have started to leverage MLLMs to solve VQA problems, as they have been trained on massive data. Q-Bench~\cite{wu2023qbench} proves that MLLMs can address preliminary low-level visual tasks. Q-Align~\cite{wu2023qalign} proposes to train a MLLM using text-defined levels (\eg, fine, poor) and achieves SOTA results on IQA and VQA tasks. Though many VQA models have been proposed, they are originally designed for natural videos and do not consider text-video alignment. 


\begin{figure*}[t]
  \centering
  \setlength{\abovecaptionskip}{2mm}
  \subfloat[Word cloud of selected prompts.]{
    \includegraphics[width=0.41\linewidth]{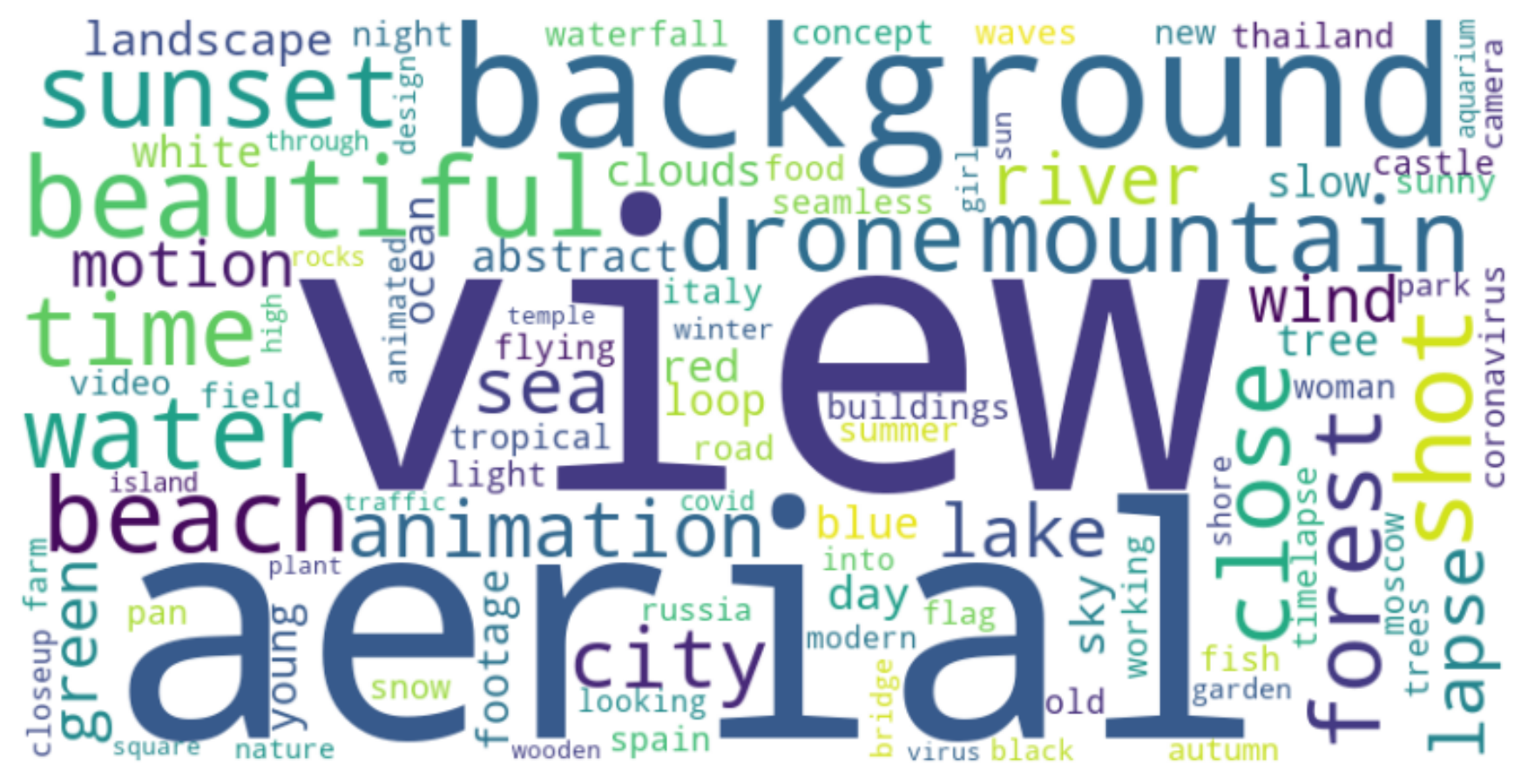}
    \label{fig:wordcloud}}
  \hfill
  \subfloat[The raw/Z-score MOS distribution.]{
    \includegraphics[width=0.28\linewidth]{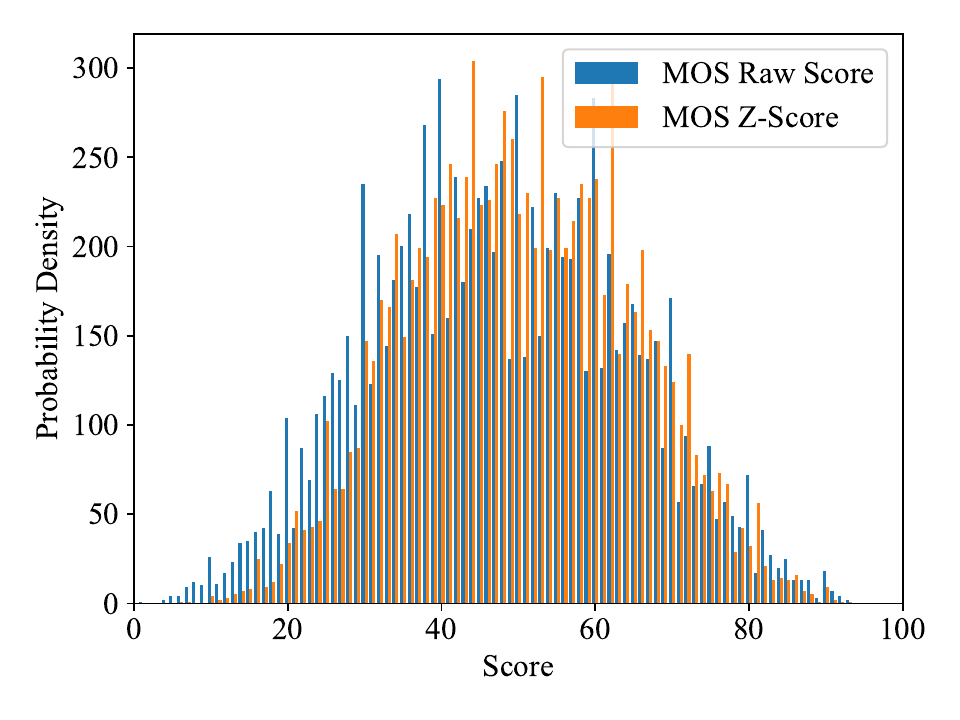}
    \label{fig:mosz}}
  \hfill
  \subfloat[The MOSz distributions of 10 T2V models.]{
    \includegraphics[width=0.27\linewidth]{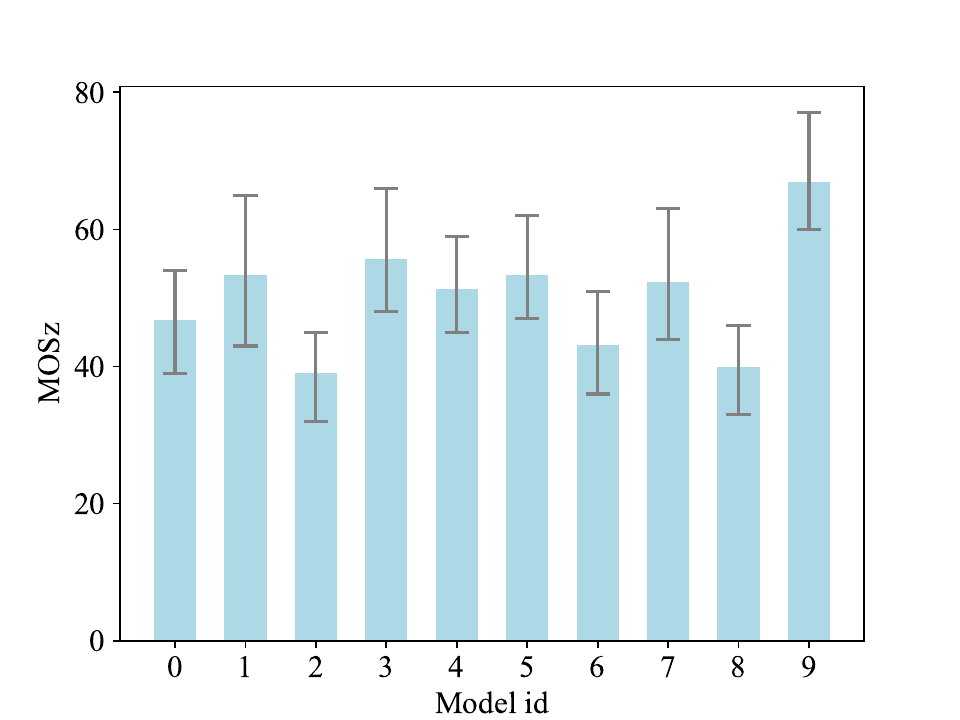}
    \label{fig:10mosz}}
  \caption{(a) The word cloud of the prompts used in T2VQA-DB. (b) The distribution of the raw/Z-score MOS. (c) Z-score MOS distributions of 10 T2V models. The model IDs represent sequentially Text2Video-Zero, AnimateDiff, Tune-a-video$^{(1)}$, VidRD, ModelScope, VideoFusion, LVDM, Show-1, Tune-a-video$^{(2)}$, LaVie. }
  \label{fig:dataset}
  \vspace{-3mm}
\end{figure*}

\begin{figure}[t]
    \centering
    \setlength{\abovecaptionskip}{0.1cm}
    \includegraphics[width=0.9\linewidth]{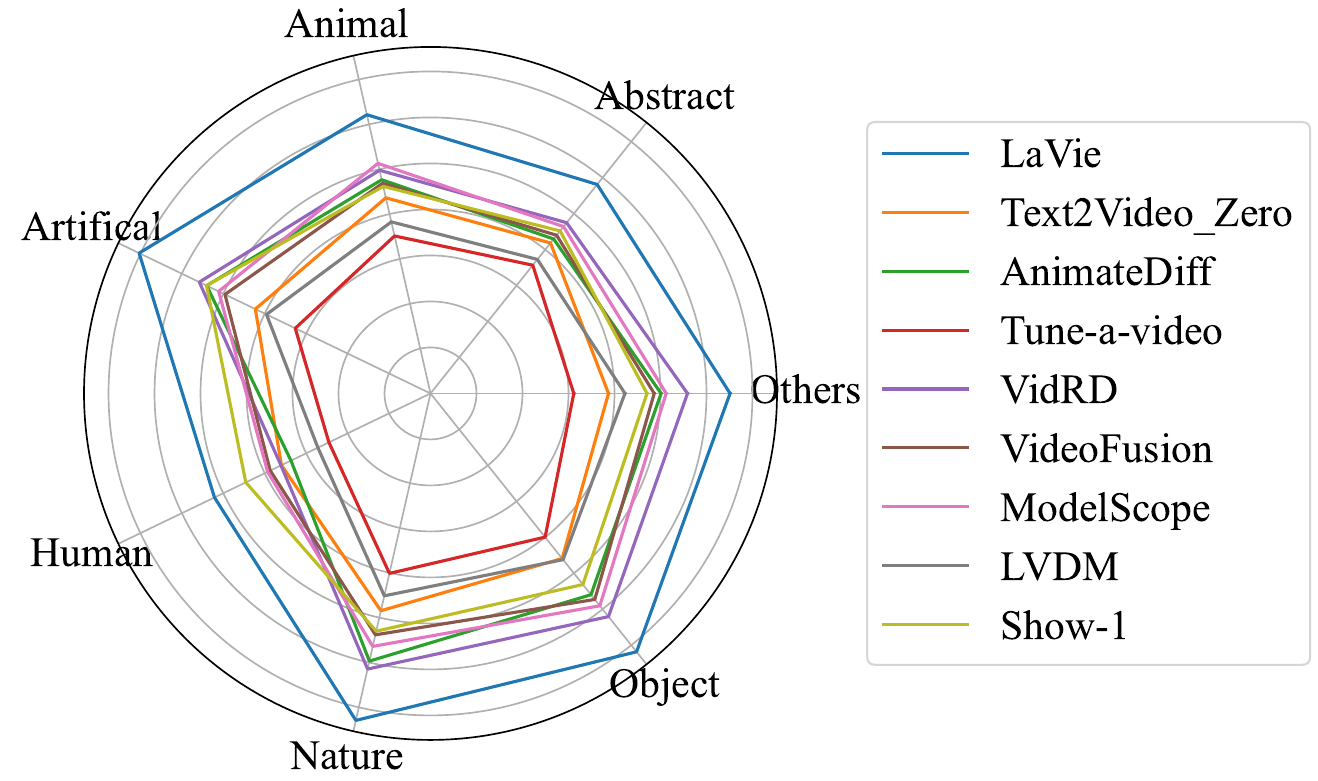}
    \caption{Comparison of models performance on different prompt types.}
    \label{fig:radio}
    \vspace{-3mm}
\end{figure}

\subsection{Text-to-video Generation}

T2V generation refers to a form of conditional video generation, where text descriptions are used as conditioning inputs to generate high-fidelity videos. 
A common practice is to extend pre-trained Text-to-Image (T2I) models with temporal modules. CogVideo~\cite{hong2022cogvideo} is based on CogView2~\cite{ding2022cogview2} and proposes a multi-frame-rate hierarchical training strategy to better align text-video clips. Make-a-video~\cite{singer2022make} adds effective spatial-temporal modules on a diffusion-based T2I model (\ie, DALLE-2~\cite{ramesh2022hierarchical}). VideoFusion~\cite{luo2023videofusion} also leverages the DALLE-2 and presents a decomposed diffusion process. LVDM~\cite{he2022latent}, Text2Video-Zero~\cite{khachatryan2023text2video}, Tune-A-Video~\cite{wu2023tune}, AnimateDiff~\cite{guo2023animatediff}, Video LDM~\cite{blattmann2023align}, MagicVideo\cite{zhou2022magicvideo}, ModelScope~\cite{wang2023modelscope}, and VidRD~\cite{gu2023reuse} are models that inherit the success of Stable Diffusion (SD) ~\cite{rombach2022high} for video generation. Show-1~\cite{zhang2023show} integrates both pixel-based and latent-based text-to-Video Diffusion Models (VDMs). LaVie~\cite{wang2023lavie} extends the original transformer block in SD to a spatio-temporal transformer. Recently, OpenAI releases Sora~\cite{videoworldsimulators2024}, a T2V model that is capable of generating 60s high-fidelity videos, considered as a game changer in T2V generation.  

\begin{figure*}[t]
    \centering
    \setlength{\abovecaptionskip}{2mm}
    \includegraphics[width=0.9\linewidth]{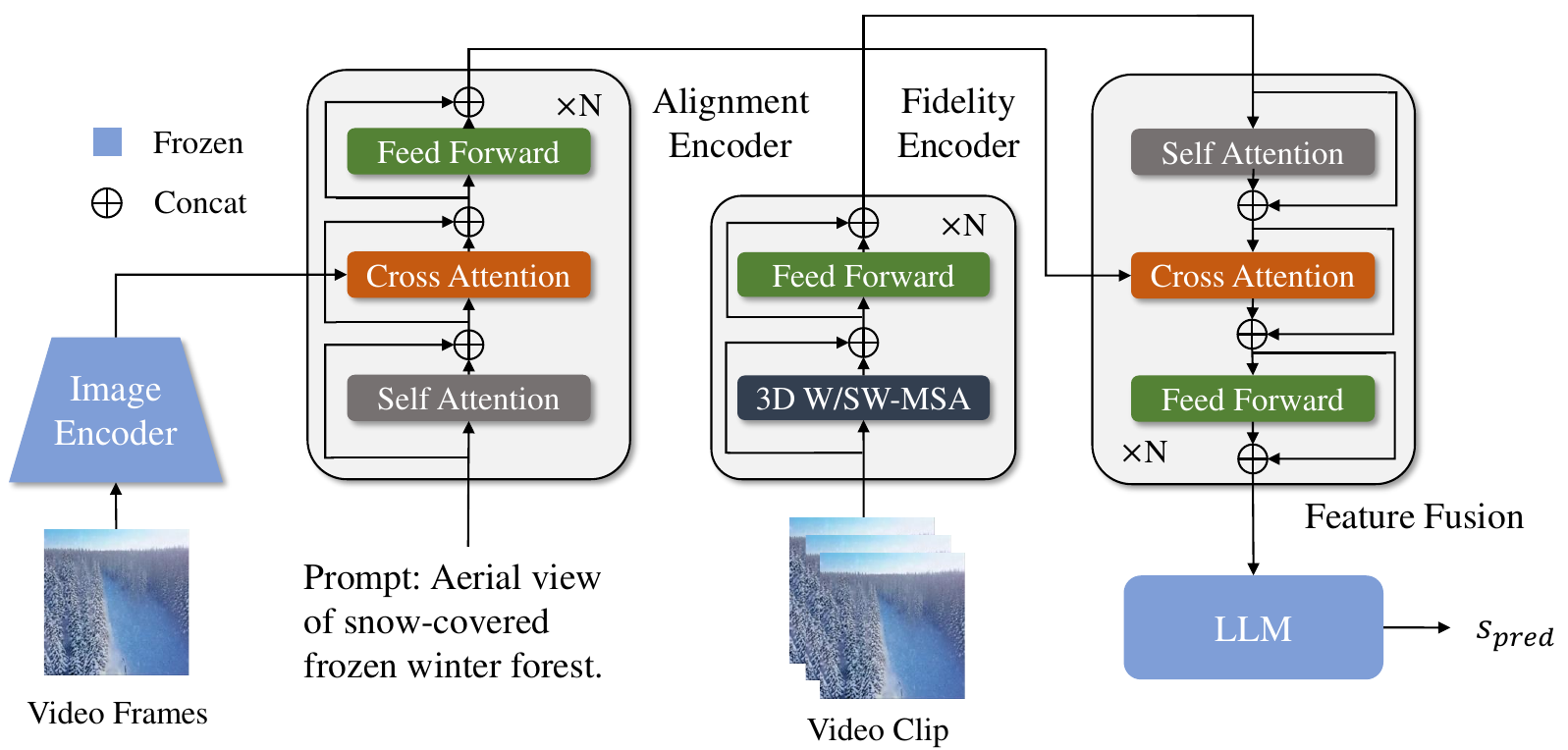}
    \caption{Overview framework of T2VQA. Features from text-video alignment and video fidelity perspectives are extracted. After a cross-attention based fusion, an LLM is utilized for regression.}
    \label{fig:model}
    \vspace{-0.4cm}
\end{figure*}

\section{Subjective-Aligned Text-to-Video Dataset}

Tab.~\ref{tab:dataset} shows most existing T2V datasets have relatively small number of videos, which is not sufficient to represent the diverse performance of T2V generation models. Consequently, we propose a Text-to-Video Quality Assessment DataBase, named T2VQA-DB, including 10,000 videos generated by 9 different T2V models. 1,000 prompts are used and 27 subjects are invited to obtain the MOS of each video. In this section, we will describe the establishment of T2VQA-DB and the subjective experiment.

\subsection{Data Preparation}

\subsubsection{Prompt Selection.}

To guarantee the diversity of the dataset, the prompts used for T2V generation should cover as many aspects as possible. Following~\cite{xu2024imagereward}, we use the same graph-based algorithm from~\cite{su2022selective} for prompt selection. We first randomly sample 1 million prompts from WebVid-10M~\cite{Bain21}, which contains 10 million video-text pairs scraped from the stock footage sites. Each prompt is encoded to a vector representation by Sentence-BERT~\cite{reimers2019sentence}. The graph-based algorithm integrates them into $k$ groups according to cosine distance. $k$ is a hyper-parameter and we set $k=100$, resulting roughly 10,000 prompts in each group. Finally, we randomly sample 10 prompts in each group, forming the 1,000 prompts in T2VQA-DB. Fig.~\ref{fig:wordcloud} shows the word cloud of the collected prompts.

\subsubsection{Video Generation.}

We use in total 9 different models for video generation, including Text2Video-Zero~\cite{khachatryan2023text2video}, AnimateDiff~\cite{guo2023animatediff}, Tune-a-video~\cite{wu2023tune}, VidRD~\cite{gu2023reuse}, VideoFusion~\cite{luo2023videofusion}, ModelScope~\cite{wang2023modelscope}, LVDM~\cite{he2022latent}, Show-1~\cite{zhang2023show}, and LaVie~\cite{wang2023lavie}. For Tune-a-video, we utilize two different pre-trained weights, resulting in a total of 10 models for generation. Compared to other T2V datasets, we utilize current advanced T2V generation models as much as possible, making T2VQA-DB more representative. Since the default resolution, video length, and frame rate are different in each model, we unify the video format as $512\times512$, 16 frames, and 4fps. 

\subsection{Subjective Study}

To obtain the MOS of each video, we invite 27 subjects to score the perceptual quality of each video. The subjects are asked to score mainly from two aspects, in terms of text-video alignment and video fidelity. Text-video alignment refers to how the generated video content matches the text description. Video fidelity refers to degrees of distortion, saturation, motion consistency, and content rationality. The subjects use a slider ranging from 0 to 100 to give the final score of each video. 
After having the raw MOS of each subject, we conduct normalization to avoid inter-subject scoring differences as Z-score MOS (MOSz). That is:


\begin{equation}
  MOSz_{i} = \frac{1}{N}\sum_{j=1}^{N}Res(\frac{r_{ij} - \mu_{j}}{\sigma_{j}}).
  \label{eq:mosz}
\end{equation}
$i$ and $j$ refer to the index of videos and subjects. $r$ is the raw score, and $\mu_j = \frac{1}{M}\sum_{i=1}^{M}r_{ij}$, $\sigma_j = \sqrt{\frac{1}{M-1}\sum_{i=1}^{M}(r_{ij}-\mu_{j})^2}$. $M$ is the number of videos scored by each subject. $N$ is the number of scores on one video. $Res(\cdot)$ is the rescaling function, converting the distribution of Z-scores into a mean of 50 and a standard deviation of 16.6. Fig.~\ref{fig:mosz} shows the distributions of raw MOS and Z-score MOS.

\subsection{Dataset Analysis}

We conduct comprehensive experiments and analysis on T2VQA-DB. We first investigate each model's performance in T2VQA-DB. The visualization of the Z-score MOS distributions of 10 models is shown in Fig.~\ref{fig:10mosz}. LaVie has the highest average MOS of 66.9, while two Tune-a-Video models have the lowest of 39.1 and 39.9. The reason for the poor performance of Tune-a-Video is mainly the low inter-frame consistency, as shown in Fig.~\ref{fig:ex3}. 

    

Based on the prompt contents, we classify the collected prompts into 6 categories, including nature, human, artificial, animal, object, and abstract. The ones that cannot be categorized into the 6 classes are labeled as ``others''. After classification, we have 327 prompts for nature, 119 for human, 196 for artificial, 121 for animal, 66 for object, 135 for abstract, and 36 for others. Subsequently, we compare the models' performance over different prompt types. As shown in Fig.~\ref{fig:radio}, LaVie outperforms the other models on all types of prompts. Tune-a-video has the worst performance, which is consistent with the analysis in Fig.~\ref{fig:10mosz}. LVDM has the second-worst performance except in the ``others'' type. The performance of the rest models has negligible differences. Fig.~\ref{fig:radio} also shows prompts labeled as human have the worst performance in all models. The reason could be that human faces and actions require more sophisticated modeling compared with other categories. Some models also show a preference for object and nature types of prompts to a small extent. 

\begin{table*}[tb]
\setlength{\abovecaptionskip}{2mm}
\centering
\renewcommand{\arraystretch}{1.2}
\caption{Performance of the SOTA models and T2VQA. The best model is highlighted in each column. [\textbf{Bold}: the best].}
\begin{tabular}{ c|c|cccc|cc|cc|cc}
\hline
\multirow{2}{*}{Type} & \multirow{2}{*}{Models} & \multicolumn{4}{c|}{T2VQA-DB Validation} & \multicolumn{2}{c|}{ECTV Testing} & \multicolumn{2}{c|}{TVGE Testing} & \multicolumn{2}{c}{Sora Testing}        \\ \cline{3-12} 
                                               & & SRCC $\uparrow$  & PLCC $\uparrow$ & KRCC $\uparrow$ & RMSE $\downarrow$ & SRCC $\uparrow$  & PLCC $\uparrow$ & SRCC $\uparrow$  & PLCC $\uparrow$ & SRCC $\uparrow$  & PLCC $\uparrow$\\ \hline
 \multirow{5}{*}{zero-shot} &
 CLIPSim~\cite{radford2021learning}                     & 0.1047 & 0.1277 & 0.0702 & 21.683 & 0.3111 & 0.3175 & 0.2765 & 0.3013 & 0.2116 & 0.1538\\
                      & BLIP~\cite{li2022blip}                      & 0.1659 & 0.1860 & 0.1112 & 18.373 & 0.3841 & 0.3877 & 0.3437 & 0.3696 & 0.2126 & 0.1038\\
 & ImageReward~\cite{xu2024imagereward}                    & 0.1875 & 0.2121 & 0.1266 & 18.243 & 0.5192 & 0.5107 & 0.4740 & 0.4879 & 0.0992 & 0.0415\\  
  & ViCLIP~\cite{wang2023internvid}                    & 0.1162 & 0.1449 & 0.0781 & 21.655 & 0.4130 & 0.4105 & 0.3429 & 0.3691 & 0.2567 & 0.1844\\
  & UMTScore~\cite{liu2024fetv}                    & 0.0676 & 0.0721 & 0.0453 & 22.559 & 0.2406 & 0.2214 & 0.2155 & 0.2199 & 0.2594 & 0.0840\\ \hline

\multirow{4}{*}{finetuned} & SimpleVQA~\cite{sun2022deep}                & 0.6275 & 0.6338 & 0.4466 & 11.163 & 0.1959 & 0.1391 & 0.3297 & 0.2219 & 0.0340 & 0.2344\\
                      & BVQA~\cite{li2022blindly}                & 0.7390 & 0.7486 & 0.5487 & 15.645 & 0.2929 & 0.2829 & 0.4283 & 0.4229 & 0.4235 & 0.2489\\
                      & FAST-VQA~\cite{wu2022fast}                & 0.7173 & 0.7295 & 0.5303 & 10.595 & 0.3101 & 0.3246 & 0.3335 & 0.3466 & 0.4301 & 0.2369\\
                      & DOVER~\cite{wu2023dover}                & 0.7609 & 0.7693 & 0.5704 & 9.8072 & 0.3118 & 0.3289 & 0.3943 & 0.4058 & 0.4421 & 0.2689\\
                      & Q-Align~\cite{wu2023qalign} & 0.7601 & 0.7768 & 0.5860 & 10.9911 & 0.4356 & 0.4401 & 0.5317 & 0.5209  & 0.3109 & 0.1997\\\hline
 \textbf{Ours} & \textbf{T2VQA}                     & \textbf{0.7965}        &  \textbf{0.8066}     & \textbf{0.6058}      & \textbf{9.0221} & \textbf{0.5694} & \textbf{0.5823} & \textbf{0.6898} & \textbf{0.6912} & \textbf{0.6485} & \textbf{0.3124} \\ \hline
\end{tabular}
\label{tab:performance}
\vspace{-3mm}
\end{table*}

\section{Subjective-Aligned T2V Metric}

Based on T2VQA-DB, we propose a novel model that leverages transformer-based architecture for Text-to-video Quality Assessment (T2VQA). The model dissolves the task into two perspectives, in terms of text-video alignment and video fidelity. After feature extraction and feature fusion, an LLM is used for quality regression. Fig.~\ref{fig:model} shows the overview of the architecture of T2VQA. We will introduce the detailed design of T2VQA below.

\subsection{Text-video Alignment Encoder}

Text-video alignment refers to the conformity between the video content and the text description. CLIP~\cite{radford2021learning} and BLIP~\cite{li2022blip} have strong abilities for zero-shot text-image matching. ~\cite{wu2021godiva} proposes CLIPSim, which uses CLIP to calculate the similarities between text and each frame of the video and then take the average value. However, former works~\cite{chivileva2023measuring, huang2024vbench, liu2024fetv,liu2023evalcrafter} and results in Tab.~\ref{tab:performance} show that simply using CLIP or BLIP has a low correlation with the authentic subjective scores. Following works of MLLMs~\cite{li2023blip2, instructblip,ye2023mplug}, we use the pre-trained BLIP image encoder as the video frame encoder. We freeze the weights of the image encoder that encodes each frame separately. We use the BLIP text encoder as the alignment encoder. The alignment encoder takes the encoded text and video frame as inputs. They interact through the cross-attention module and the encoder eventually outputs a feature representing the text-frame similarity. We concatenate features from each text-frame pair. Given a set of $N$ video frames $\{v_i \in \mathbb{R}^{3 \times H \times W}\}_{i=1}^{N}$ and the text prompt $t$, we have:

\begin{equation}
    f_{b} = cat(\{\mathrm{BLIP}(t, v_i)\}_{i=1}^{N}),
\end{equation}

where $cat$ is short for concatenation.

\subsection{Video Fidelity Encoder}

Video fidelity refers to the perception of distortion from spatial and temporal domains. In the spatial domain, common distortion types include blurriness, noises, low/high contrast, \etc Temporal distortions include jitter, stall, motion blur, \etc In this perspective, the task can be seen as a common VQA task. Swin-T~\cite{liu2022video} has been proven for its excellent ability in various VQA tasks~\cite{wu2022fast, wu2023dover}. By using 3D-shifted window-based multi-head self-attention (SW-MSA), Swin-T has a strong ability to analyze videos from spatial and temporal domains. Therefore, we utilize Swin-T as the backbone of the fidelity encoder to extract features that represent video fidelity. Given a video clip $v\in \mathbb{R}^{3\times N \times H \times W}$, we have:

\begin{equation}
    f_{s} = \mathrm{SWIN}(v).
\end{equation}

\subsection{Feature Fusion}

Inspired by BLIP-2~\cite{li2023blip2} and InstructBLIP~\cite{instructblip}, after having the features from perspectives of text-video alignment and video fidelity, we design a transformer-based fusion module to fuse those two features. The module includes $N$ blocks with self-attention, cross-attention, and feed-forward layers in each block. The fidelity feature $f_s$ first goes through self-attention layers, and then it interacts with the alignment feature $f_b$ in cross-attention layers (and every other transformer block). The fusion module helps the model to unify the features from two perspectives to have a more comprehensive understanding of the video characteristics. We initialize the fusion module using $\mathrm{BERT}_{base}$~\cite{kenton2019bert}.

\subsection{Quality Regression}

LLMs have been proven to be competitive on quality assessment tasks~\cite{wu2023qinstruct,wu2023qalign, zhang2023q, wu2024towards, zhang2024lmm}. Inspired by them, we also utilize an LLM as the quality regression module in T2VQA. We first design a text instruction prompt, \ie, ``\textit{Please rate the quality of this video.}'', to guide the LLM. The encoded instruction and the fused feature are concatenated as the input of the LLM. Following~\cite{liqe, wu2023qalign}, we supervise the LLM to output one among five ITU-standard~\cite{bt2002methodology} levels  (\textit{bad}, \textit{poor}, \textit{fair}, \textit{good}, and \textit{excellent}) to represent the quality of the videos, denoted as \texttt{<level>}. We assign them weights of $1-5$ in order. Since the logit at \texttt{<level>} in LLM is the probability distribution of all tokens, it can be used to represent how the LLM predicts the quality of the video. Therefore, a softmax for each token is calculated and multiplied by its weight. We have the final predicted score as:

\begin{equation}
    s_{pred} = \sum_{i=1}^5 i \times softmax(\lambda_i) = \sum_{i=1}^5 i \times \frac{e^{\lambda_i}}{\sum_{j=1}^5 e^{\lambda_j}},
\end{equation}
where $\lambda_i$ is the probability distribution of the i-th \texttt{<level>} token.

\section{Experiments and Results}

\begin{figure*}[tb]
  \centering
  \setlength{\abovecaptionskip}{2mm}
  \subfloat[BLIP~\cite{li2022blip}]{
    \includegraphics[width=0.23\linewidth]{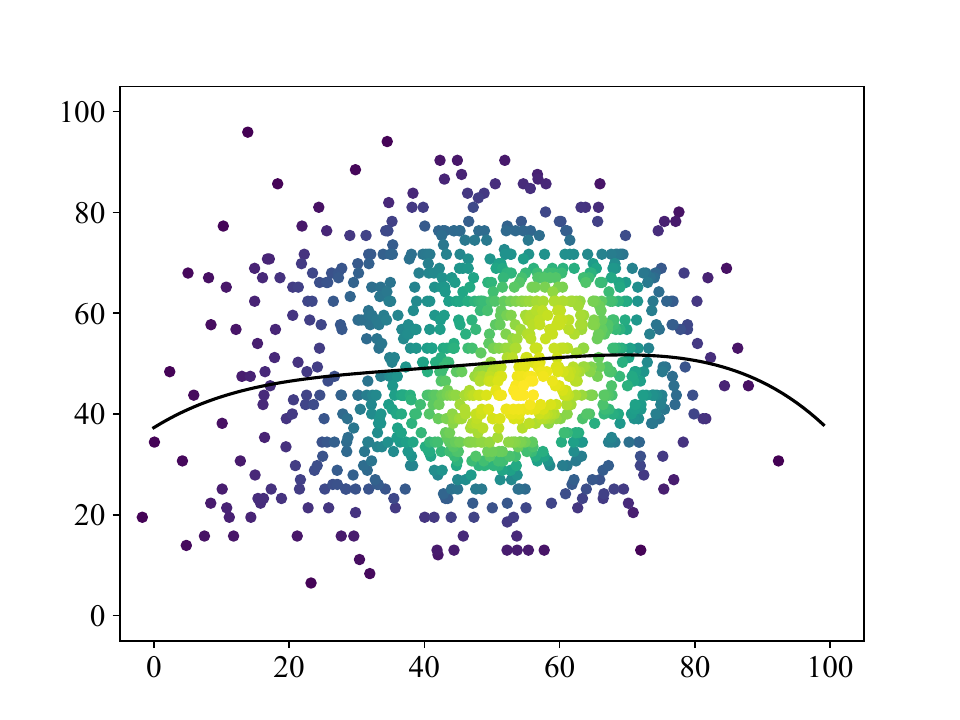}
    \label{fig:blip}}
  \hfill
  \subfloat[ImageReward~\cite{xu2024imagereward}]{
    \includegraphics[width=0.23\linewidth]{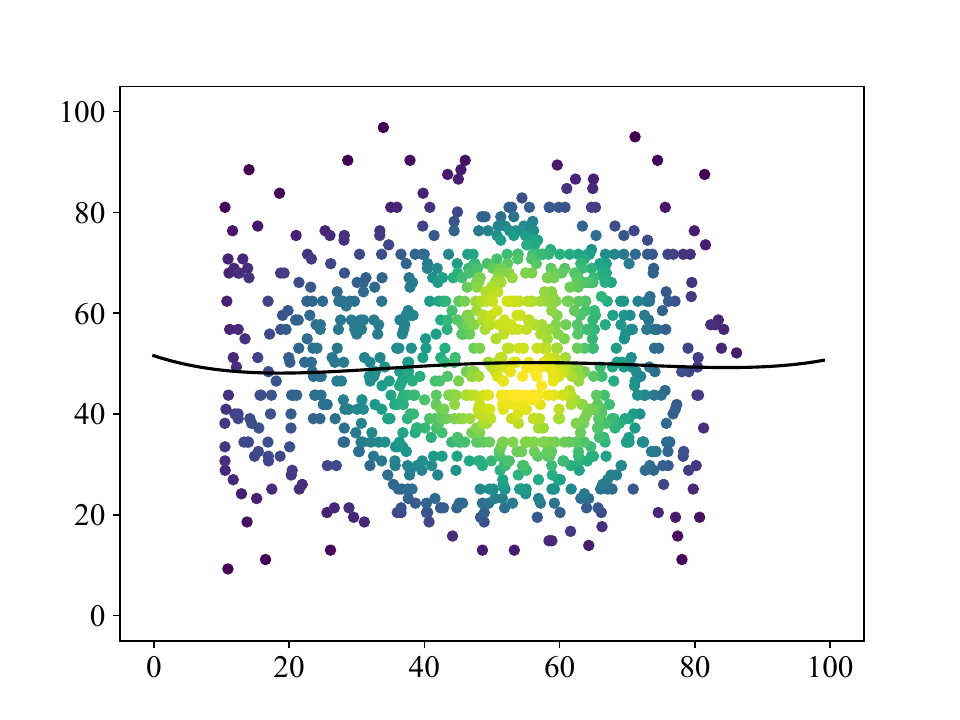}
    \label{fig:imgrwd}}
  \hfill
  \subfloat[SimpleVQA~\cite{sun2022deep}]{
    \includegraphics[width=0.23\linewidth]{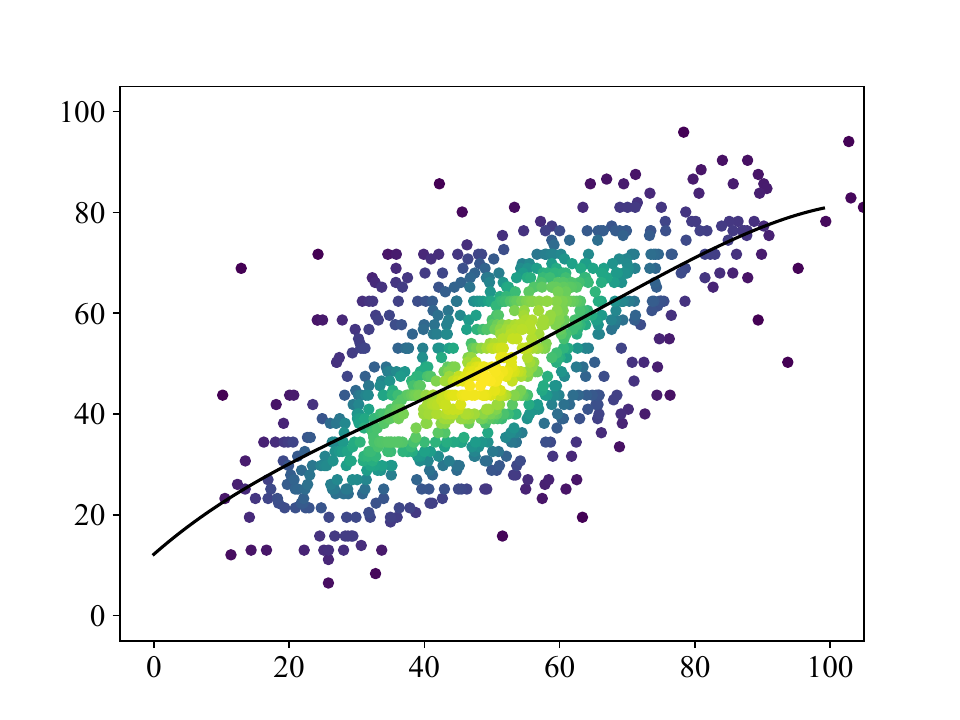}
    \label{fig:simple}}
  \hfill
  \subfloat[BVQA~\cite{li2022blindly}]{
    \includegraphics[width=0.23\linewidth]{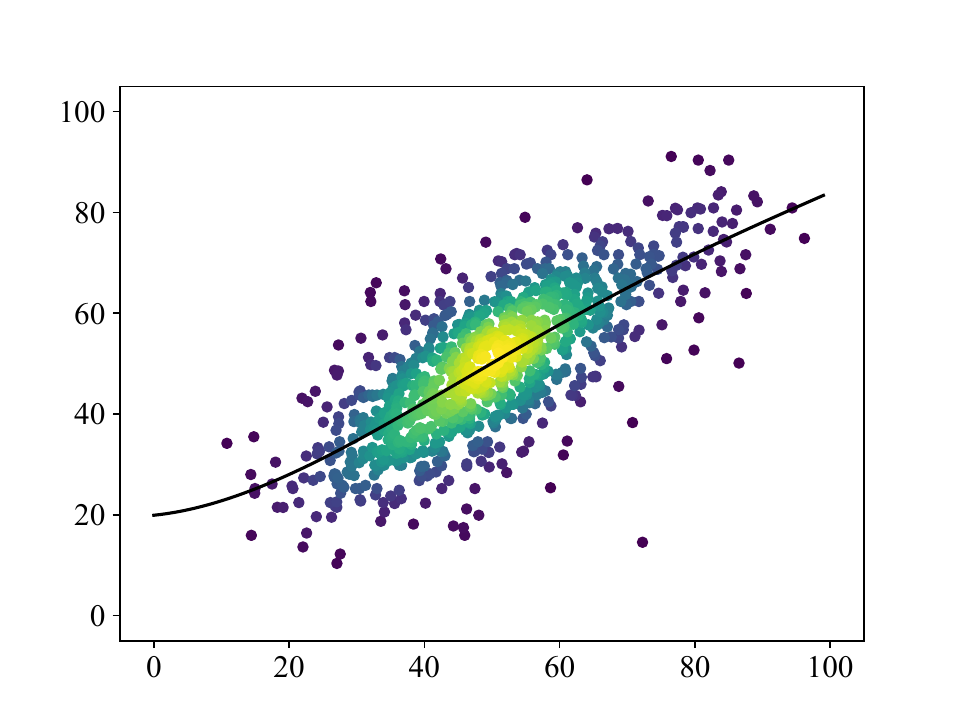}
    \label{fig:bvqa}}
  \\ \vspace{-3mm}
  \subfloat[FAST-VQA~\cite{wu2022fast}]{
    \includegraphics[width=0.23\linewidth]{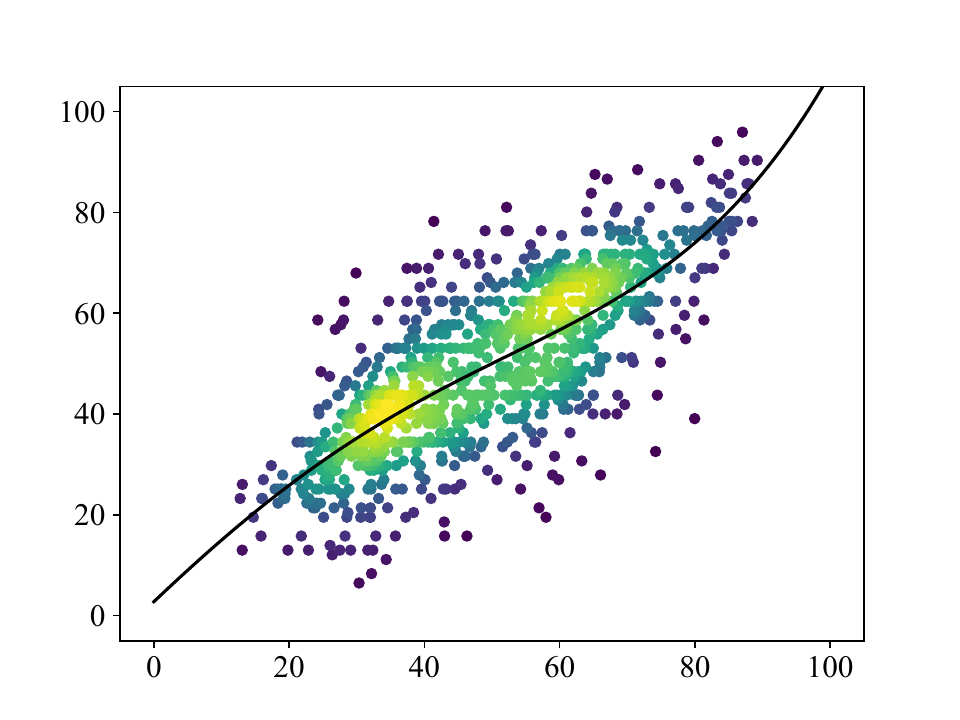}
    \label{fig:fast}}
  \hfill
  \subfloat[DOVER~\cite{wu2023dover}]{
    \includegraphics[width=0.23\linewidth]{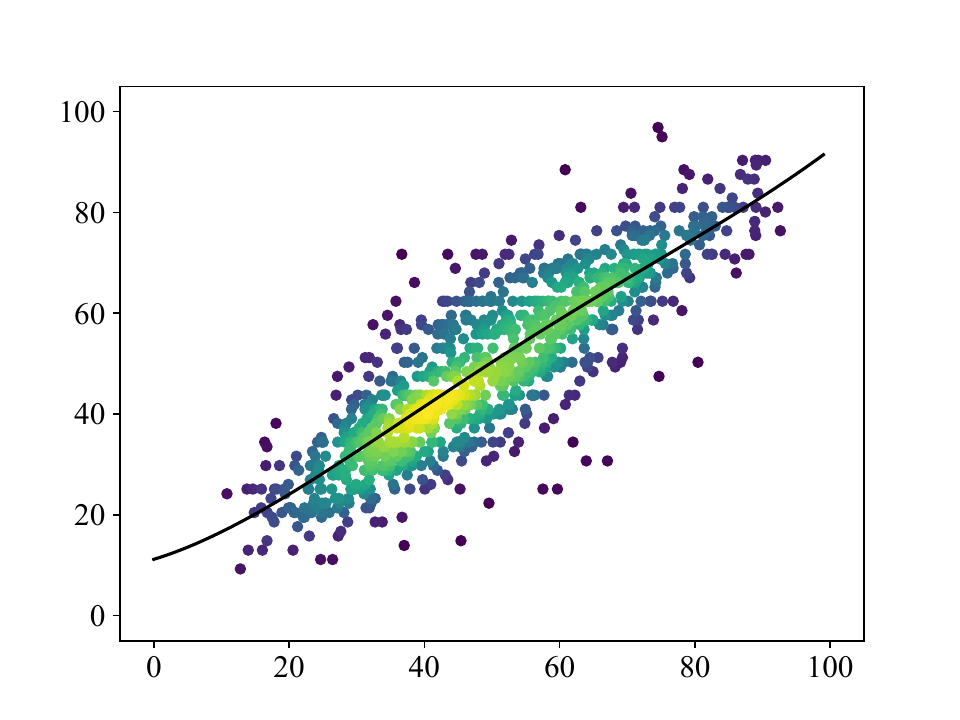}
    \label{fig:dover}}
  \hfill
  \subfloat[Q-Align~\cite{wu2023qalign}]{
    \includegraphics[width=0.23\linewidth]{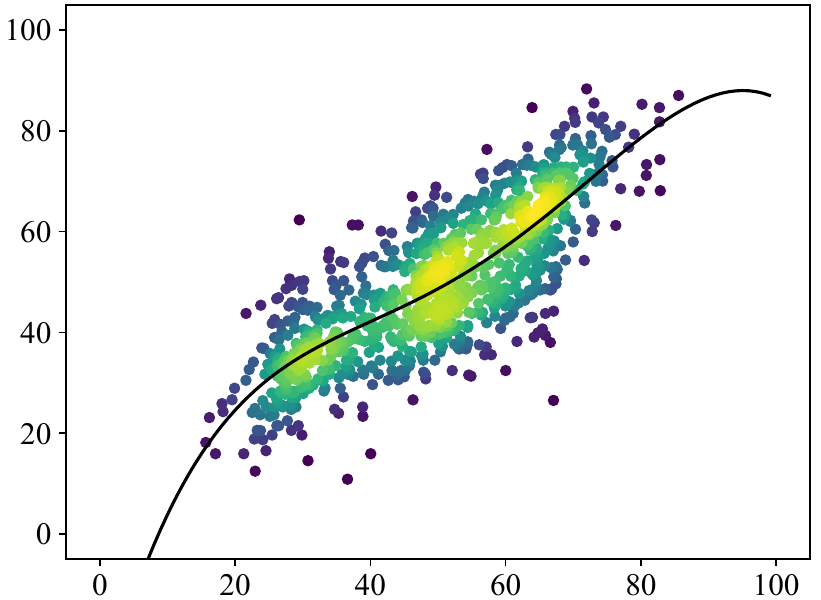}
    \label{fig:qalign}}
  \hfill
  \subfloat[T2VQA]{
    \includegraphics[width=0.23\linewidth]{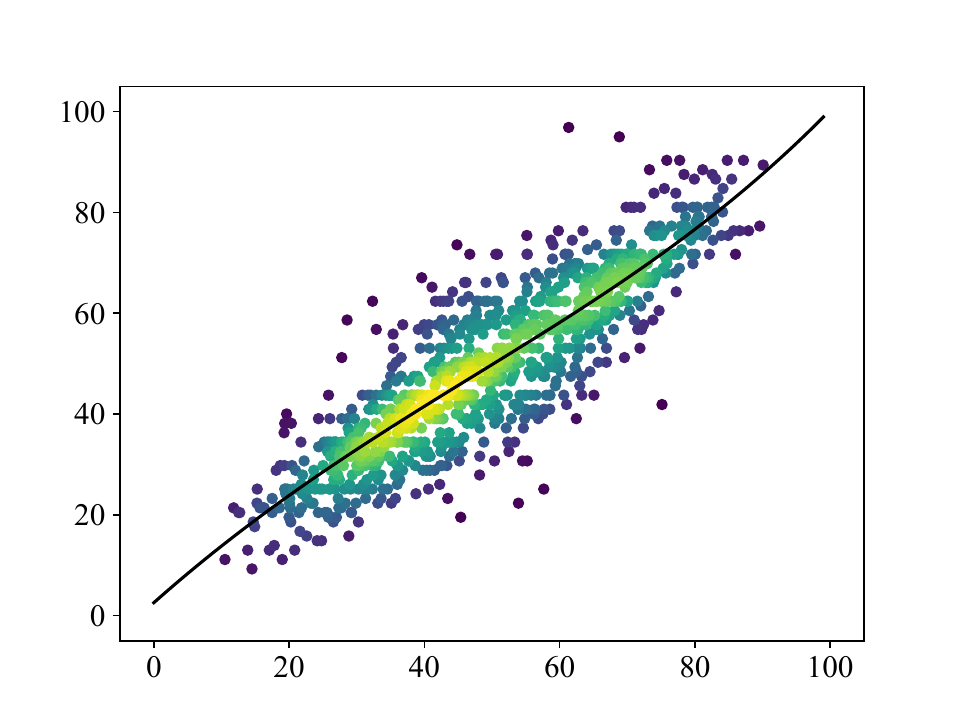}
    \label{fig:t2vqa}}
  \caption{Scatter plots of the predicted scores vs. MOSs. The curves are obtained by a four-order polynomial nonlinear fitting. The brightness of scatter points from dark to bright means density from low to high. }
  \label{fig:distribution}
  \vspace{-3mm}
\end{figure*}

\subsection{Implement Details}

\subsubsection{Train-test splitting.}

When training and testing on T2VQA-DB, we follow the common practice of dataset splitting by leaving out 80\% for training, and 20\% for testing. To eliminate the bias in one single split, we randomly split the dataset 10 times, and use the average results for performance comparison. 

\subsubsection{Training Settings.}

We utilize the large model of BLIP for the extraction of text-video alignment feature. We initialize the fidelity encoder using Swin-T pre-trained on Kinetics-400~\cite{kay2017kinetics} dataset. For the LLM, we use the 7B model of Vicuna v1.5~\cite{zheng2024judging}, which is fine-tuned from Llama 2~\cite{touvron2023llama}. It is worth noting that both BLIP image encoder and LLM are frozen during training. 

During training and testing, we first uniformly sample 8 frames out of an input video, and then we resize them to $224\times224$. We use Adam optimizer initialized by a learning rate of $1e-5$. The learning rate decays under a cosine scheduler from 1 to 0. We train T2VQA for 30 epochs under a batch size of 4 on a server with one NVIDIA GeForce RTX 4090. 

\subsubsection{Loss Function.}

Following~\cite{wu2022fast}, we use differentiable Pearson Linear Correlation Coefficient (PLCC) and rank loss as loss functions. PLCC is a common criterion used for evaluating the correlation between sequences, while rank loss is introduced to help the model distinguish the relative quality of videos better. The final loss function is defined as:

\begin{equation}
    L = L_{plcc} + \lambda \cdot L_{rank}.
\end{equation}
$\lambda$ is a hyper-parameter for balancing, and is set to $0.3$ in training.

\subsubsection{Evaluation Metrics.}

Along with PLCC, we include Spearman’s Rank-order Correlation Coefficient (SRCC), Kendall’s Rank-order Correlation Coefficient (KRCC), and Root Mean Square Error (RMSE) as performance criteria. 
Better models should have larger SRCC, KRCC, and PLCC scores, but conversely for RMSE. Before calculating PLCC, we follow~\cite{video2000final} to map the objective score to the subject score using a four-parameter logistic function.

\subsection{Performance Comparison}

\subsubsection{Reference Algorithms.}

We use CLIPSim~\cite{wu2021godiva}, BLIP~\cite{li2022blip}, ImageReward~\cite{xu2024imagereward}, ViCLIP~\cite{wang2023internvid}, UMTScore~\cite{liu2024fetv}, SimpleVQA~\cite{sun2022deep}, BVQA~\cite{li2022blindly}, FAST-VQA~\cite{wu2022fast}, DOVER~\cite{wu2023dover}, and Q-Align~\cite{wu2023qalign} as the reference algorithms. Q-Align is considered the SOTA VQA method to date. CLIPSim averages the similarity values between the text and each video frame. We adopt the same operation on BLIP and ImageReward to tune IQA metrics into VQA metrics. ViCLIP and UMTScore are metrics designed for measuring text-video alignment. SimpleVQA, BVQA, FAST-VQA,  DOVER, and Q-Align are models designed for general VQA tasks. We use the pre-trained weights of CLIPSim, BLIP, and ImageReward for zero-shot testing, since they are originally designed for generated content. SimpleVQA, BVQA, FAST-VQA, DOVER, and Q-Align are originally designed for UGC videos, so we finetune them on T2VQA-DB. All results are averaged after ten-fold splitting. 

\subsubsection{Results Analysis.}

The first four columns of number in Tab.~\ref{tab:performance} show the performance comparison between T2VQA and other SOTA models on T2VQA-DB. Results show that T2VQA performs best in SRCC, surpassing Q-Align by 4.79\% in SRCC and 3.84\% in PLCC. The zero-shot models all have relatively low scores. They either only consider the text-video alignment or don't analyze the temporal domain information within video frames. The VQA models have higher scores, indicating that video fidelity heavily affects the assessment of text-generated video quality. However, a single perspective from video fidelity cannot address the problem properly, as there are circumstances where a high-fidelity video is generated but does not match the prompt.

Fig.~\ref{fig:distribution} shows scatter plots between the predicted scores and the Z-Score MOSs of BLIP~\cite{li2022blip}, ImageReward~\cite{xu2024imagereward}, ViCLIP~\cite{wang2023internvid}, SimpleVQA~\cite{sun2022deep}, BVQA~\cite{li2022blindly}, FAST-VQA~\cite{wu2022fast}, DOVER~\cite{wu2023dover}, and T2VQA. The figure uses 1,000 videos, which are randomly sampled from the testing set in one split of T2VQA-DB. A better model should have a fitted curve close to the diagonal and have less dispersed scatter points. As shown in Fig.~\ref{fig:distribution}, T2VQA also outperforms the others. 

\subsubsection{Cross-dataset Validation.}

ECTV~\cite{liu2023evalcrafter} and TVGE~\cite{wu2024better} are the only two T2V datasets that have released their subjective scores. We randomly sample 800 videos from each dataset for cross-dataset validation. Besides, Sora~\cite{videoworldsimulators2024} has been considered the SOTA T2V generation model.
We collect the videos generated by Sora from its official website to validate the generalization of T2VQA and other models on high-fidelity videos.
We invite 20 annotators to score the quality of each Sora video. T2VQA and other reference VQA models are trained on T2VQA-DB and tested on the sampled ECTV, TVGE, and Sora videos. We report the SRCC and PLCC between the models predictions and the ground truth MOSs. The results are listed in the six columns on the right side of Tab.~\ref{tab:performance}. 


Experimental results show that T2VQA has the best generalization ability among all models. Noticed that there is a performance drop between the validation on T2VQA-DB and the testing on other datasets. That's because videos from the other datasets have the attributes of high resolution, high frame rate, and long length, which current open-sourced T2V models are not able to generate. We will include videos with high fidelity in T2VQA-DB in future work.  

\subsubsection{Qualitative Analysis.}

We also conduct a qualitative analysis on three examples with good, fair, and poor quality. 
We use SimpleVQA~\cite{sun2022deep}, BVQA~\cite{li2022blindly}, FAST-VQA~\cite{wu2022fast}, DOVER~\cite{wu2023dover}, and T2VQA to predict their quality. Fig.~\ref{fig:qualititive} presents the prompts, video frames, and model predictions on the three examples, and Tab.~\ref{tab:example} lists the models' predictions and MOSs.   Results show that T2VQA has more subjective-aligned predictions. Fig.~\ref{fig:ex1} shows an example with a relatively high score. The scene in the video matches the description in the prompt well, and the video frames are clear and consistent. Fig.~\ref{fig:ex2} is a medium-level example. Though the video matches the prompt basically, the video frames suffer from blurriness, which is reflected in the MOS and T2VQA's prediction. Fig.~\ref{fig:ex3} shows the worst case. The video fails to accurately present the description in the prompt. Besides, it loses consistency between video frames. Although it has a high definition in each frame separately, it still has low scores in both MOS and T2VQA's prediction.

\subsection{Ablation Studies}

\begin{table}[t]
\setlength{\abovecaptionskip}{2mm}
\renewcommand{\arraystretch}{1.2}
    \centering
    \caption{T2VQA and other models predictions on 3 example videos. Colored numbers represent the distance between predictions and the ground truths. {\color{red}Red}: the best. }
    \begin{tabular}{c|c|c|c}
        \hline
        Model & Prompt 1 & Prompt 2 & Prompt 3 \\
        \hline
        SimpleVQA~\cite{sun2022deep} & $72.89_{\color{green}-5.33}$ & $61.59_{\color{green}+3.88}$ & $49.95_{\color{green}+17.39}$ \\
        BVQA~\cite{li2022blindly} & $73.12_{\color{green}-5.1}$ & $60.37_{\color{green}+2.66}$ & $45.68_{\color{green}+13.12}$ \\
        FAST-VQA~\cite{wu2022fast} & $85.28_{\color{green}+7.06}$ & $44.19_{\color{green}-13.52}$ & $38.91_{\color{green}+6.35}$ \\
        DOVER~\cite{wu2023dover} & $89.11_{\color{green}+10.89}$ & $51.93_{\color{green}-5.78}$ & $36.66_{\color{green}+4.1}$ \\
        \textbf{T2VQA(Ours)} & $81.61_{\color{red}+3.39}$ & $57.11_{\color{red}-0.6}$ & $29.07_{\color{red}-3.49}$ \\
        \hline
        MOS (gt) & 78.22 & 57.71 & 32.56 \\
         \hline
    \end{tabular}
    
    \label{tab:example}
    \vspace{-3mm}
\end{table}

\begin{figure}[t]
    \setlength{\abovecaptionskip}{2mm}
    \centering
    \subfloat[Prompt: Castle ruins on the hill in the middle of a beautiful landscape.]{
                \includegraphics[width=\linewidth]{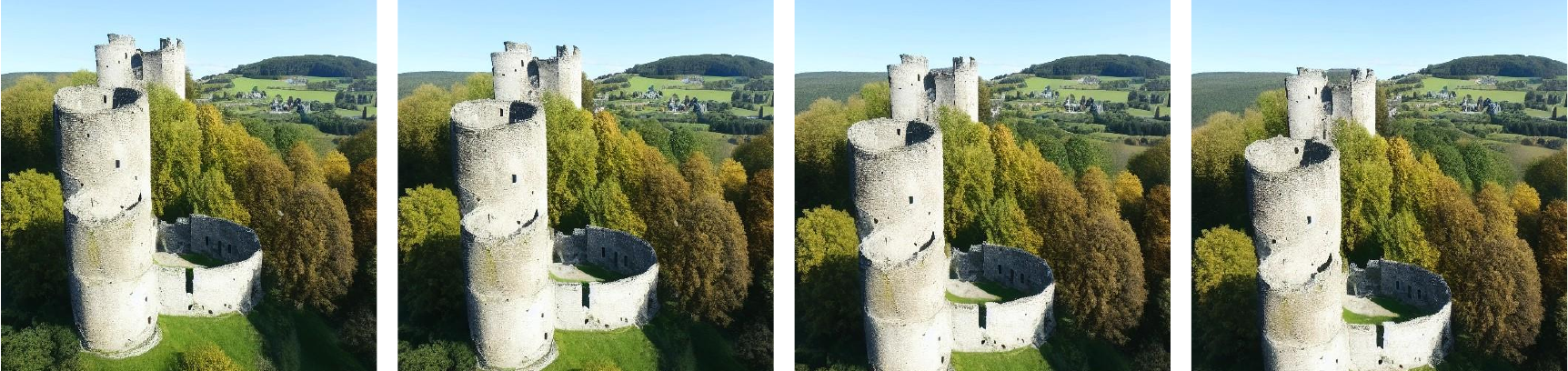}
                \label{fig:ex1}
                }\\ \vspace{-3mm}
    \subfloat[Prompt: Underwater world with different fishes, corals, and stones.]{
                \includegraphics[width=\linewidth]{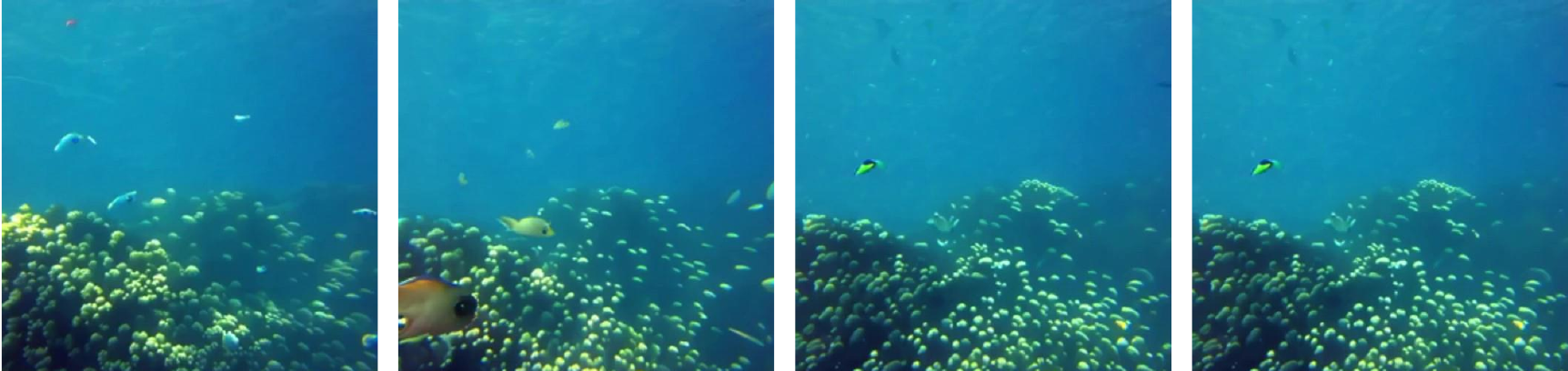}
                \label{fig:ex2}
                }\\ \vspace{-3mm}
    \subfloat[Prompt: In a hot cast-iron cauldron, the cook pours oil to fry the meat (liver).]{
                \includegraphics[width=\linewidth]{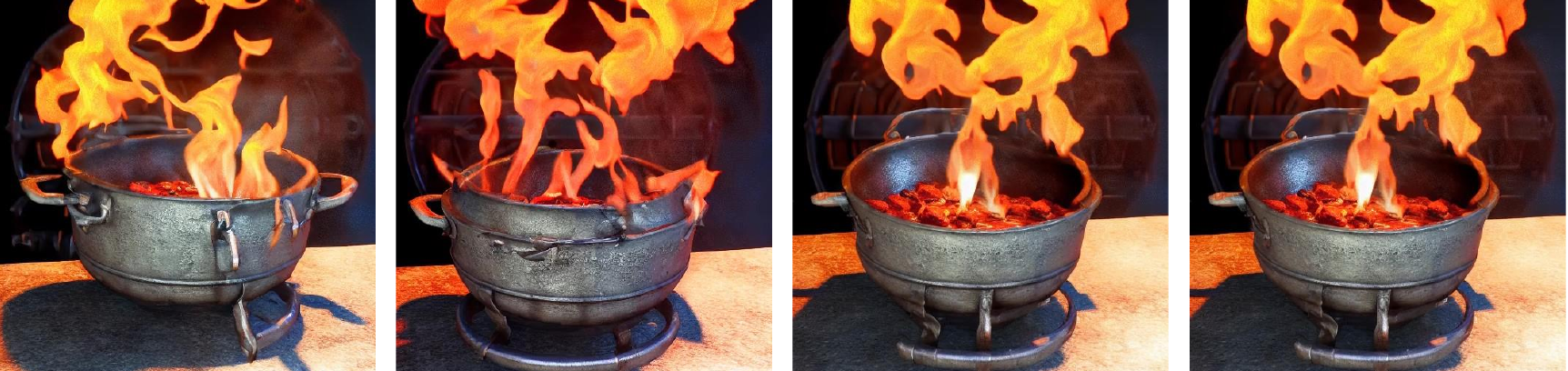}
                \label{fig:ex3}
                }
    \caption{Three videos with good, fair, and poor quality.}
    \label{fig:qualititive}
    \vspace{-3mm}
\end{figure}

To validate the effectiveness of each module in T2VQA, we conduct thorough ablation studies, including the alignment and fidelity encoder, the fusion module, and the regression module. Results are listed in Tab.~\ref{tab:ablation}. All results are averaged after 10-fold splitting.

\subsubsection{Alignment Encoder.}

T2VQA utilizes the BLIP image encoder and text encoder as the alignment encoder. CLIP is another widely used text-image encoder. It has a similar structure and ability to measure text-video alignment as BLIP. We replace the BLIP image and text encoder in T2VQA with CLIP to determine which one has the better performance. 

Experimental results show that using CLIP as the alignment encoder suffers from severe performance degradation. The reason could be that the image and the text are encoded separately in CLIP. While in BLIP, the encoded image and the text features interact in the cross-attention module.  

\begin{table}[t]
\centering
\setlength{\abovecaptionskip}{2mm}
\renewcommand{\arraystretch}{1.2}
\caption{Results of ablation studies. [Keys: \textbf{Bold}: the best].}
\begin{tabular*}{\linewidth}{@{\extracolsep{\fill}} ccccc}
\hline
 \multirow{2}{*}{Models} & \multicolumn{4}{c}{Validation}         \\ \cline{2-5} 
                                              & SRCC $\uparrow$  & PLCC $\uparrow$ & KRCC $\uparrow$ & RMSE $\downarrow$ \\ \hline
  T2VQA-CLIP                     & 0.7296 & 0.7347 & 0.5385 & 10.5141 \\
    T2VQA-resnet                      & 0.7610 & 0.7730 & 0.5715 & 9.8152  \\
   T2VQA-cat                    & 0.7734 & 0.7854 & 0.5839 & 9.4034 \\  
  T2VQA-linear                & 0.7808 & 0.7919 & 0.5891 & 9.3011 \\
                        T2VQA-nonlinear                & 0.7850 & 0.7983 & 0.5954 & 9.1755   \\
 \textbf{T2VQA(Ours)}                     & \textbf{0.7965}        &  \textbf{0.8066}     & \textbf{0.6058}      & \textbf{9.0221}   \\ \hline
\end{tabular*}
\label{tab:ablation}
\vspace{-3mm}
\end{table}

\subsubsection{Fidelity Encoder.}
In T2VQA, we use the Swin-T as the backbone of the fidelity encoder. To investigate the effectiveness of the transformer-based architecture, we conduct the control experiment by using the convolution-based 3D ResNet~\cite{hara3dcnns} as the fidelity encoder. The results show that Swin-T has a superior performance to ResNet, validating its effectiveness.  



\subsubsection{Fusion and Regression Modules.}
Besides, we compare our cross-attention fusion strategy with the simple concatenation fusion. The latter is the simplest yet most commonly used fusion strategy. In T2VQA, we take advantage of the strong ability of an LLM for quality regression. We test the commonly used linear regression and non-linear regression to compare with the LLM regression in T2VQA. For linear regression, we use two full-connected layers with 128  neurons in the first layer and 1 neuron in the second. For non-linear regression, we use two 1D convolution blocks with kernel size set to 1. We also set the channel number to 128 in the first block and 1 in the second. 

Results in Tab.~\ref{tab:ablation} show that T2VQA achieves the best performance in all evaluating metrics, indicating its effectiveness. The models using concatenation,  linear regression, and non-linear regression have similar performance, yet they are all inferior to T2VQA, indicating that the cross-attention fusion and LLM have achieved non-negligible improvement to the model.



\section{Conclusion}

In conclusion, in this paper, we are dedicated to giving a subjective-aligned prediction of the quality of a text-generated video. For that purpose, we establish a T2V dataset with the largest scale, named T2VQA-DB. The dataset includes 10,000 videos generated by 9 advanced T2V models. We also conduct a subjective study to obtain the MOSs on the overall video quality. Based on T2VQA-DB, we propose a novel transformer-based model for text-to-video quality assessment, named T2VQA. The model extracts features of a video from text-video alignment and video fidelity perspectives respectively. After fusing the features, an LLM is utilized to regress the final prediction. The experimental results indicate that T2VQA is effective in evaluating the quality of text-generated videos.

\begin{acks}

This work was supported in part by the National Natural Science Foundation of China under Grant 62301310, 62225112, and U1908210, and in part by the Shanghai Pujiang Program under Grant 22PJ1406800, and in part by Sichuan Science and Technology Program under Grant 2024NSFSC1426.

\end{acks}

\bibliographystyle{ACM-Reference-Format}
\balance
\bibliography{sample-base}










\end{document}